\documentclass[conference]{IEEEtran}
\IEEEoverridecommandlockouts
\usepackage{cite}
\usepackage{amsmath,amssymb,amsfonts}
\usepackage{algorithmic}
\usepackage{graphicx}
\usepackage{textcomp}
\usepackage{xcolor}
\usepackage{amsmath}
\usepackage{amssymb}
\usepackage{multirow}
\usepackage{subfigure}
\usepackage{graphicx}

\graphicspath{{./fig/}}

\def\BibTeX{{\rm B\kern-.05em{\sc i\kern-.025em b}\kern-.08em
    T\kern-.1667em\lower.7ex\hbox{E}\kern-.125emX}}
\def\etal{{\em et al.}}
\begin{document}

\title{Low-Resolution Face Recognition via Adaptable Instance-Relation Distillation
\thanks{This work was partially supported by grants from the Pioneer R\&D Program of Zhejiang Province (2024C01024), and Open Research Project of the State Key Laboratory of Media Convergence and Communication, Communication University of China (SKLMCC2022KF004). Shiming Ge is the corresponding author (geshiming@iie.ac.cn).}
}
\author{\IEEEauthorblockN{1\textsuperscript{st} Ruixin Shi, 2\textsuperscript{nd} Weijia Guo, 3\textsuperscript{rd} Shiming Ge}
\IEEEauthorblockA{\textit{Institute of Information Engineering, Chinese Academy of Sciences}\\
\textit{School of Cyber Security, University of Chinese Academy of Sciences}\\
Beijing, China \\
shiruixin@iie.ac.cn, guoweijia@iie.ac.cn, geshiming@iie.ac.cn}

}

\maketitle

\begin{abstract}
Low-resolution face recognition is a challenging task due to the missing of informative details. Recent approaches based on knowledge distillation have proven that high-resolution clues can well guide low-resolution face recognition via proper knowledge transfer. However, due to the distribution difference between training and testing faces, the learned models often suffer from poor adaptability. 
To address that, we split the knowledge transfer process into distillation and adaptation steps, and propose an adaptable instance-relation distillation approach to facilitate low-resolution face recognition. 
In the approach, the student distills knowledge from high-resolution teacher in both instance level and relation level, providing sufficient cross-resolution knowledge transfer. 
Then, the learned student can be adaptable to recognize low-resolution faces with adaptive batch normalization in inference. 
In this manner, the capability of recovering missing details of familiar low-resolution faces can be effectively enhanced, leading to a better knowledge transfer.
Extensive experiments on low-resolution face recognition clearly demonstrate the effectiveness and adaptability of our approach.
\end{abstract}


\section{Introduction}
Low-resolution face recognition plays a vital role in many practical applications, such as verifying face pairs in video surveillance and classifying face identities in automatic driving. Although, deep models have proven success in face recognition applications~\cite{deng2019arcface} on public benchmark~\cite{LFWTech}, accuracy would decrease when directly applying them to recognize low-resolution ones in practical scenarios. On the one hand, there is a resolution gap challenge: low-resolution faces contain much fewer informative details and recognizable-related discriminative features.
On the other hand, there is a data gap challenge: the distribution difference between the well-organized training faces and testing low-resolution ones cannot be neglected in the open-set settings.
Another intuitive way is to train a new model on massive low-resolution faces from scratch, which is very time and labor-consuming. 
Consequently, fully leveraging pre-trained off-the-shelf deep models to facilitate low-resolution face recognition tasks is an effective yet economical solution, which can be divided into two types.
The hallucination-based methods~\cite{grm2019face, kuo2022towards} aim to reconstruct high-resolution faces before recognizing them. However, these methods impose non-ignorable computational burden.
In contrast, the embedding-based methods~\cite{chai2023recognizability,zha2019tcn,wang2022two} directly recognize low-resolution faces leveraging transferred knowledge.
The knowledge here can be grouped into two levels.
Instance-level knowledge transfer draws teacher and student representation space closer in a sample-to-sample way~\cite{zhao2022dkd}. However, this transferring may be limited and insufficient, which cannot preserve inter-sample relations well.
Relation-level knowledge transfer attempts to transfer structural relations between samples~\cite{park2019relational}, where relation matters a lot in recognition tasks~\cite{liu2022coupleface}. In this way, high-order knowledge can be mined and transferred.
Also, contrastive learning has been introduced to gain better performance~\cite{tian2020iclr,zhu2021complementary}.
Such distillation-based methods can well guide low-resolution face recognition via proper knowledge transfer of high-resolution clues~\cite{wang2023low}. 
However, due to the difference of data distribution between training and testing faces, the learned models often suffer from poor adaptability. Moreover, the gap issues mentioned above are still not well addressed.
\begin{figure}[t]
\centering
\centerline{\includegraphics[width=1.0\linewidth]{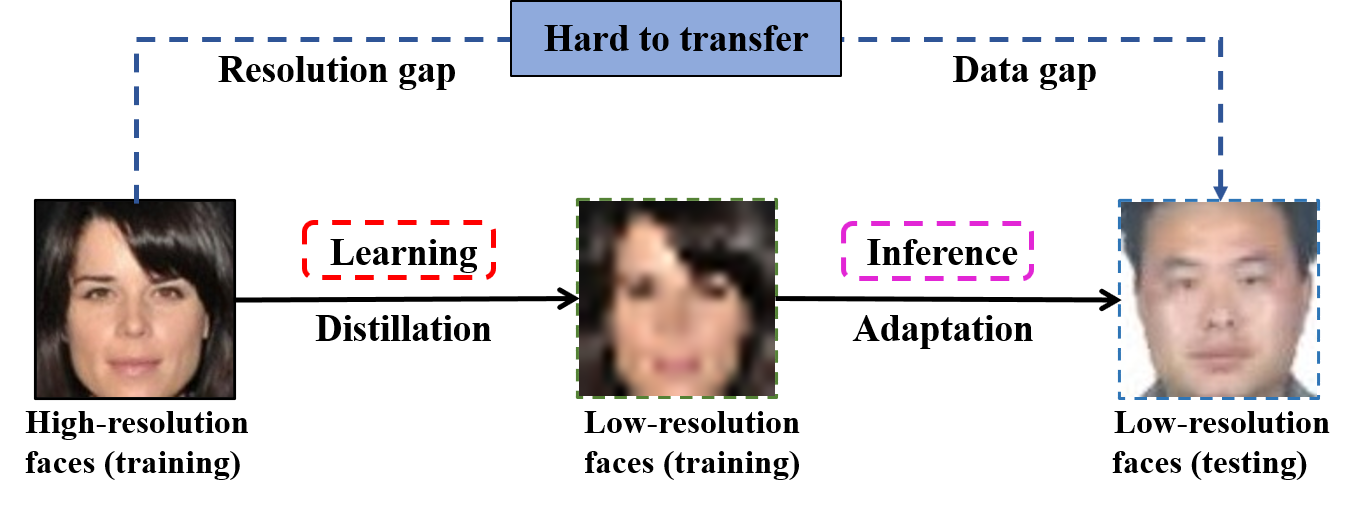}}
\caption{Our motivation. Direct knowledge transfer from high-resolution faces to low-resolution ones is hard due to resolution and data gaps. Thus, we conduct distillation in learning and perform adaptation in inference to enable sufficient and adaptable transfer.}
\label{motivation}
\end{figure}

To address that, we split the knowledge transfer process into distillation and adaptation steps, and propose an adaptable instance-relation distillation (AIRD) approach to facilitate low-resolution face recognition, as shown in Fig.~\ref{motivation}.
The teacher contains rich knowledge of high-resolution faces, while the student learns to generate discriminative features without losing accuracy and preserve the network performance.
Our method first performs instance-relation distillation in a teach-student learning manner, which aims to reduce the large resolution gap by making the student mimic the behaviors of a well-trained teacher in the representation space, providing sufficient cross-resolution knowledge transfer.
To do so, we propose instance-level distillation module (IlD) which distills well-constructed decoupled smooth probability constraints between the predictions point-to-point, and relation-level distillation module (RlD) which distills contrastive relational structure knowledge between the teacher and student representation space.
By combining them, our approach not only can align the embedding space of student and teacher well, but also can maximize the inter-class discriminative ability and the intra-class compactness for low-resolution face recognition.
Our instance-relation distillation applies a similar concept with CRCD~\cite{zhu2021complementary} and DKD~\cite{zhao2022dkd}, but has two main differences: i) The motivation is different. Our method is designed for cross-resolution knowledge transfer in low-resolution face recognition task. In addition to distilling a large network into a smaller one, we need to maximize the inter-class distance while minimizing the intra-class distance, considering low-resolution faces contain much fewer recognizable-related details, as shown in Fig~\ref{framework}; ii) The distillation algorithm is very different. Unlike CRCD~\cite{zhu2021complementary} which transfers relation knowledge from the same domain without considering sample difference and DKD~\cite{zhao2022dkd} which transfers knowledge point-to-point, our method focuses on the harder low-resolution samples by pre-selecting positive and negative pairs offline, leveraging the idea of hard example mining because low-resolution faces always have small inter- and intra-class distance. Also, a more specific facial classifier replaces the original softmax layer. 

While the student model has learned enough knowledge to close the resolution gap, it can still be improved when recognizing faces in unknown environments since the datasets' distribution can be quite different. 
In light of this, we propose an inference adaptation method to improve transfer ability of the learned model in recognizing real-world low-resolution faces by introducing an online domain adaptive batch normalization (FaceBN) technique. 
Adaptive batch normalization is often used for effective test-time adaptation by updating the parameters of batch normalization layers with testing data. Practical face recognition task like surveillance face searching needs to extract features from a set of low-resolution gallery faces to match a probe face, where FaceBN is a suitable and effective solution because data distribution is inconsistent between faces captured in surveillance and those used for training.
It narrows the non-negligible data gap between a well-organized low-resolution training faces and testing faces by replacing the statistics in batch normalization layers across the student network to more closely match the actual distribution of testing faces without any additional parameters or components.
In this way, the learned student can be adaptable to recognize testing low-resolution faces effectively.

Our major contributions are as follows: 1) we propose an adaptable instance-relation distillation approach to facilitate low-resolution face recognition by combining instance-level distillation and relation-level distillation, 2) we propose an inference adaptation method to improve transfer ability of the learned model in recognizing real-world low-resolution faces, and 3) we conduct experimental evaluations to demonstrate that our proposed approach achieves state-of-the-art performance in low-resolution face recognition.

\section{Related Work}
\subsection{Low-Resolution Face Recognition}
Previous studies have illustrated that directly applying off-the-shelf deep models trained on public high-resolution datasets to recognize low-resolution ones is flawed.
Consequently, fully leveraging the prior knowledge of pre-trained high-resolution face recognition models to facilitate low-resolution face recognition tasks is a reasonable solution, which can be divided into two categories.
The hallucination-based methods reconstruct high-resolution faces from low-resolution ones before recognizing them~\cite{grm2019face,9578478}. However, hallucination methods increase computational complexity and impose a computational burden.

In contrast, the embedding-based methods directly recognize low-resolution faces based on the knowledge transferred from pre-trained high-resolution face recognition models.
Zangeneh \etal~\cite{zangeneh2020low} proposed a new coupled mapping method consisting of two DCNN branches for mapping high and low-resolution faces to non-linear transformed public space. Zha \etal \cite{zha2019tcn} proposed an end-to-end transferable coupling network in high-resolution and low-resolution domains respectively.
In addition, knowledge distillation is a teacher-student learning framework, which has been proven effective for facilitating visual applications. 
On the one hand, the instance-level knowledge transfer aims to draw the representation space of teacher and student model closer by reducing the distance of individual samples' output.~\cite{Hinton2015DistillingTK,kim2017transferring,mirzadeh2020improved, zhao2022dkd} aim to directly imitate the neural response of the last output layer of the teacher model. While~\cite{RomeroBKCGB14,heo2019comprehensive,chen2021cross} mimic the intermediate representations of teacher model to improve the learning of student model by matching original or transformed features. 
The second one is relation-level knowledge transfer which attempts to transfer structural relations between samples of outputs rather than individual outputs themselves~\cite{tung2019similarity,peng2019correlation,park2019relational,tian2019crd,zhu2021complementary}. 
Tung \etal~\cite{tung2019similarity} used pairwise activation similarities in each input mini-batch to supervise the student learning, and Park \etal~\cite{park2019relational} proposed to transfer explicit sample relations from pretrained teacher. 
Besides, contrastive learning has been introduced to mine more complex high-dimensional relations~\cite{zhu2021complementary}.
From these works, we find transferring the knowledge from high-resolution to low-resolution models is helpful and can avoid the computationally-intensive face reconstruction.

\subsection{Adaptive Batch Normalization}
Adaptive batch normalization has been extensively researched to enhance model generalization, which is often used for effective test-time adaptation by updating the parameters of batch normalization layers with test data.
Batch normalization is a typical method used to mitigate the negative impact of domain gap, which can scale and shift the value of network activations during training and is fixed in the inference phase. But data distributions can be quite different, and those statistics are always inconsistent with testing dataset domain.~\cite{DBLP:conf/iclr/LiWS0H17} propose adaptive batch normalization to increase the generalization ability of a deep neural networks. By modulating the statistics from the source domain to the target domain in all batch normalization layers across the network.~\cite{klingner2022unsupervised} presents a solution to the task of unsupervised domain adaptation of a given pre-trained semantic segmentation model without relying on any source domain representations.~\cite{DBLP:conf/iclr/Niu00WCZT23} has shown to be effective at tackling distribution shifts between training and testing data by adapting a given model on test samples.

\begin{figure*}[t]
\centering
\centerline{
\includegraphics[width=1.0\linewidth]{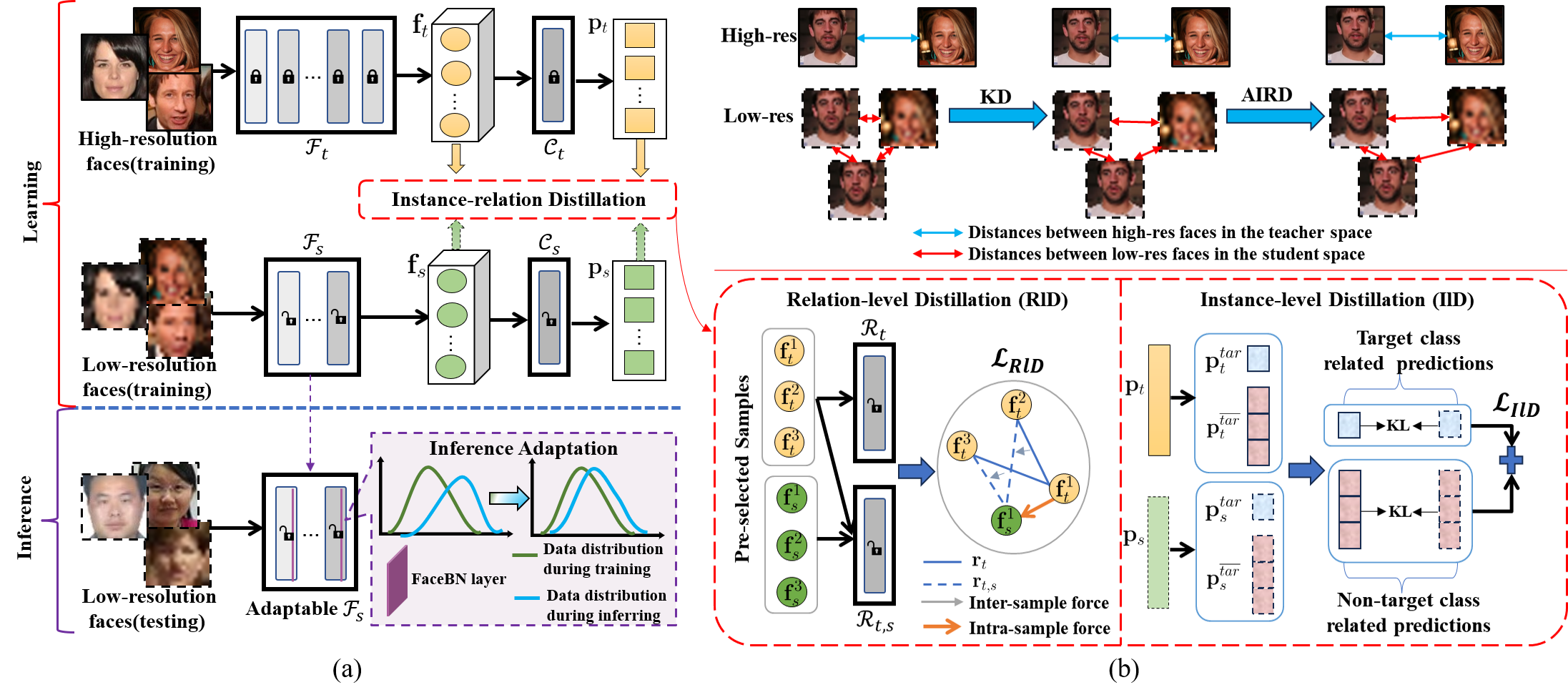}
}
\caption{Overview of our adaptable instance-relation distillation (AIRD) approach. (a) The approach performs instance-relation distillation and then the learned student model can be used to effectively recognize low-resolution faces via inference adaptation. (b) Different from traditional knowledge distillation (KD), our AIRD can maximize the inter-class distance while minimizing the intra-class distance by combining instance-level distillation and relation-level distillation.$\mathcal{F}_{t}$ and $\mathcal{C}_{t}$ are the teacher backbone and head, while $\mathcal{F}_{s}$ and $\mathcal{C}_{s}$ are the student counterparts.
$\mathbf{f}_{t}$ and $\mathbf{f}_{s}$ are the extracted features. $\mathbf{p}_{t}$ and $\mathbf{p}_{t}$ are the model predictions.
$\mathbf{r}_{t}$ and $\mathbf{r}_{t,s}$ are the extracted feature relations.}
\label{framework}
\end{figure*}
\section{Proposed Approach}

\subsection{Problem Formulation}
Our objective is learning an adaptable student model $\phi_{s}(\hat{\mathbf{x}};\mathbf{w}_s)$ with parameters $\mathbf{w}_s$ that can provide discriminative-enhanced representations for recognizing low-resolution faces $\hat{\mathbf{x}}$ in the inference.
Let $\mathbb{D}=\{(\mathbf{x}_i, \hat{\mathbf{x}}_i, {y}_i)\}_{i=1}^{|\mathbb{D}|}$ denotes the training dataset, where $\mathbf{x}_i$ denotes the $i-$th labeled high-resolution face with the identity of ${y}_i$ and $\hat{\mathbf{x}}_i$ is the corresponding down-sampled low-resolution one.
Note that the data distribution of training and testing low-resolution faces is different.
The complex teacher model $\phi_{t}(\mathbf{x};\mathbf{w}_t)$ with parameters $\mathbf{w}_t$ is in the form $\phi_{t} = (\mathcal{F}_{t},\mathcal{C}_{t})$, where $\mathcal{F}_{t}$ is a feature extraction backbone and $\mathcal{C}_{t}$ is a classification layer.
The compact student model has the same architecture $\phi_{s} = (\mathcal{F}_{s},\mathcal{C}_{s})$.
We denote $\mathbf{f}_{t}$ and $\mathbf{f}_{s}$ as extracted features, and $\mathbf{p}_{t}$ and $\mathbf{p}_{s}$ as the model predictions.

\subsection{Instance-Relation Distillation}
First, we perform an instance-relation distillation in learning. 
$\phi_{s}$ aims to close the resolution gap to approximate teacher network's ability with minimal accuracy loss, ideally having: 
\begin{equation}
\begin{aligned}
  \phi_{t}(\mathbf{x};\mathbf{w}_t) \doteq \phi_{s}( \hat{\mathbf{x}};\mathbf{w}_s).
\end{aligned}
\label{eq:formulation}
\end{equation}
The $\doteq$ means ``equivalence'' in some metrics, e.g., similarity of representations or consistence of output.
To solve it, we propose instance-level distillation module (IlD) and relation-level distillation module (RlD). By combining them, our approach can provide sufficient cross-resolution knowledge transfer, as shown in Fig.~\ref{framework}.

\textbf{RlD} aims to transfer cross-resolution contrastive relational structure knowledge. As in~\cite{liu2022coupleface}, relation matters a lot in recognition tasks, because it is important for improving the discriminative ability of feature embedding.
We formulate a new cross-resolution student relation $\mathbf{r}_{t,s}$ as relation between each $\hat{\mathbf{x}}$ and the rest $\mathbf{x}$ except its corresponding one, which can pay more attention to the complex higher-order inter-sample dependency.
And the relation in the teacher space is denoted as $\mathbf{r}_{t}$.
To model relation between features, we denote $\mathcal{R}_{t}$ and $\mathcal{R}_{t,s}$ as relation extraction networks including a linear transformation layer and a ReLU function, which can capture the complex deep representations and high-dimensional knowledge.
Here, 
\begin{equation}
\begin{gathered}
\mathbf{r}_{t}^{i,j}=\mathcal{R}_{t}(\mathbf{f}_{t}(\mathbf{x}_i),\mathbf{f}_{t}(\mathbf{x}_j)),\\
\mathbf{r}_{t,s}^{i,j}=\mathcal{R}_{t,s}(\mathbf{f}_{t}(\mathbf{x}_i),\mathbf{f}_{s}(\hat{\mathbf{x}}_j)).
\end{gathered}
\end{equation}

Inspired by contrastive learning~\cite{tian2020iclr}, we use $\mathbf{r}_{t}$ as supervision to distill $\mathbf{r}_{t,s}$ by maximizing the mutual information for contrastive objective:
\begin{equation}
\mathbb{I}\left(\mathbf{r}_{t},\mathbf{r}_{t,s} \right)=\mathbb{E}_{P\left(\mathbf{r}_{t}, \mathbf{r}_{t,s} \right)} \log \frac{P\left(\mathbf{r}_{t},\mathbf{r}_{t,s} \right)}{P\left(\mathbf{r}_{t}  \right)P\left(\mathbf{r}_{t,s}  \right)}.
\end{equation}
Here, $P\left(\mathbf{r}_{t}, \mathbf{r}_{t,s} \right)$ and $P\left(\mathbf{r}_{t} \right)$, $P\left(\mathbf{r}_{t,s}  \right)$ are the conditional joint distribution and marginal distribution.

Considering low-resolution faces always have small inter- and intra-class distance, we focus more on the harder samples, leveraging the idea of hard example mining~\cite{ShrivastavaGG16}, given the specificity of the low-resolution face recognition tasks.In a mini-batch, we select positive and negative pairs offline in advance.
The positive pairs consist of two samples with the same identity, and are sorted by the similarity scores. After embedding face pairs into a high-dimensional feature space by $\phi_{t}$, the similarity of a positive pair can be obtained by $<\phi_{t}(\mathbf{x}_i,\mathbf{x}_j)>$.
The negative pairs with different identities are selected via hard negative mining, which are selected with the largest similarities obtained by $<\phi_{t}(\mathbf{x}_i,\mathbf{x}_k)>$, where $\mathbf{x}_i$, $\mathbf{x}_k$ are from different identities. Once the similarities of positive and negative pairs are constructed, the corresponding distributions can be estimated.

To obtain a solvable loss, we maximize the mutual information by maximizing the log likelihood similarly as NCE~\cite{chen2020simple}, which is equivalent to minimizing the loss function:
\begin{equation}
\begin{aligned}
\mathcal{L}_{RlD} &=-\sum_{q(k=1)} \log h\left(\mathbf{r}^{t},\mathbf{r}^{t,s} \right) \\
&-n\sum_{q(k=0)} \log \left[1-h\left(\mathbf{r}^{t},\mathbf{r}^{t,s}\right)\right],
\end{aligned}
\end{equation}
where $q(k=1)$ acts as positive pairs, while $q(k=0)$ acts as negative pairs. The function $h$ can be any family of functions that satisfy $h:\left\{\mathbf{r}_{t},\mathbf{r}_{t,s}\right\} \rightarrow[0,1]$. 

\textbf{IlD} aims to transfer well-constructed decoupled smooth probability constraints between the predictions point-to-point. Instead of using conventional soft-target distilling methods, which are not well-designed for this task, we introduce a decoupled logits-based method to explicitly emphasizes predictions of each $\hat{\mathbf{x}}$ as in~\cite{zhao2022dkd}.
It reformulates the predictions into two parts: one is target class related which transfers the difficulty of training samples, and the other is non-target class related which contains important dark knowledge.
We define $\mathbf{p}= \left[p_1, p_2, \ldots, p_t, \ldots, p_c\right] \in \mathbb{R}^{1 \times c}$, where $p_i$ is the probability of the $i$-th class. 
$\mathbf{p}^{tar}$ is the probabilities of the target class and $\mathbf{p}^{\overline{tar}}$ are all the other probabilities of non-target classes, which can be calculated by:
\begin{equation}
\mathbf{p}^{tar}=\frac{ \exp \left(p_{tar}\right)}{\sum_{j=1}^c \exp \left(p_{j}\right)},
\mathbf{p}^{\overline{tar}}=\frac{\sum_{k=1, k \neq tar}^c \exp \left(p_{k}\right)}{\sum_{j=1}^c \exp \left(p_{j}\right)}
\end{equation}

We use KL-Divergence in IlD, and the loss function is as follows, where $\mathbf{p}^{tar}$ is the probabilities of the target class and $\mathbf{p}^{\overline{tar}}$ are all the other probabilities of non-target classes.
\begin{equation}
\begin{aligned}
\mathcal{L}_{IlD}
 &=\underbrace{\mathbf{p}_{t}^{tar} \log \left(\frac{\mathbf{p}_{t}^{tar}}{\mathbf{p}_{s}^{tar}}\right)+\mathbf{p}_{t}^{\overline{tar}} \log \left(\frac{\mathbf{p}_{t}^{\overline{tar}}}{\mathbf{p}_{s}^{\overline{tar}}}\right)}_{target} \\
 &+ \mathbf{p}_{t}^{\overline{tar}} \underbrace{\sum_{i=1, i \neq tar}^c  \hat{\mathbf{p}}^i_{t} \log \left(\frac{\hat{\mathbf{p}}^i_{t}}{\hat{\mathbf{p}}^i_{s}}\right)}_{non-target}
\end{aligned}
\label{inkd}
\end{equation}

Fig.~\ref{framework} above explains the difference between traditional distillation methods (KD) and our approach. In KD, there are cases where the distances between different classes of student space are still smaller, which is not sufficient. Our AIRD cannot only align the embedding space of student and teacher well but also maximize the inter-class distance while minimizing the intra-class distance.
In summary, the total loss of the learning stage is as follows:
\begin{equation}
\mathcal{L}=\alpha \mathcal{L}_{IlD}+\beta \mathcal{L}_{RlD}+\mathcal{L}_{Cls},
\end{equation}
where $\mathcal{L}_{Cls}$ is a general classification loss for face recognition as in the teacher model. The $\alpha$ and $\beta$ are weighting hyper-parameters to balance the loss terms. We set $\alpha=1$ and $\beta=2$ in the evaluation.
\subsection{Inference Adaptation}
While the student model has learn enough knowledge to close the resolution gap between high-resolution faces and their corresponding down-scaled ones, it can still be improved when recognizing faces in unknown environment.
Next, we conducts inference adaptation, which aims to improve the transfer ability of the learned student by adapting it to the testing faces.
We denote $p_{dl}(\hat{\mathbf{x}})$ and $p_{rl}(\hat{\mathbf{x}})$ as the data distribution of low-resolution faces during training and inferring.
The key is to eliminate the data distribution discrepancy between training and testing faces. 
By performing normalization to $p_{dl}$, the covariant shift can be removed, ideally having:
\begin{equation}
\begin{aligned}
 p_{dl}(\hat{\mathbf{x}}) p_{dl}\left(\frac{\hat{\mathbf{x}}-E_{dl}(\hat{\mathbf{x}})}{\sqrt{V_{dl}(\hat{\mathbf{x}})}}\right) \approx p_{rl}(\hat{\mathbf{x}}) p_{rl}\left(\frac{\hat{\mathbf{x}}-E_{rl}(\hat{\mathbf{x}})}{\sqrt{V_{rl}(\hat{\mathbf{x}})}}\right),
\end{aligned}
\label{eq:formulation2}
\end{equation}
where $E_{dl}$ and $V_{dl}$ represent the expectation and variance in the training data, and $E_{rl}$ and $V_{rl}$ represent the expectation and variance in the testing data.

To achieve this, we introduce FaceBN, an online domain adaptive batch normalization technique, into the distilled student model.
It replaces the parameters in all batch normalization layers across the student with the mixture of re-estimated statistics of testing faces over batches and the original statistics obtained during training without any additional components. The recalculate mean and variance are:
\begin{equation}
\begin{gathered}
\mu_B=\gamma \cdot \mu_B+(1-\gamma) \cdot \frac{1}{M} \sum_{i=1}^M \hat{\mathbf{x}}_i; \\
\sigma_B^2=\gamma \cdot \sigma_B^2+(1-\gamma) \cdot \frac{1}{M} \sum_{i=1}^M\left(\hat{\mathbf{x}}_i-\mu_B\right)^2,
\end{gathered}
\end{equation}
where $\mu_B$ and $\sigma_B^2$ denote the mean and variance value, $M$ is mini-batch size, and $\gamma $ is the momentum parameter which is set to 0.1.

\section{EXPERIMENTS}\label{sec:majhead}
To verify the effectiveness and adaptability of our approach AIRD on low-resolution face recognition, we conduct experiments on both verification and identification tasks. We use high-resolution CASIA-WebFace~\cite{yi2014learning} and its downsampled version as training dataset.
Then we evaluate the recognition accuracy on LFW~\cite{LFWTech}, Age-DB~\cite{moschoglou2017agedb} and UCCS ~\cite{uccs}.
 CASIA-WebFace~\cite{yi2014learning} is a large-scale dataset for face recognition containing 10,575 subjects and 494,414 images, which was built using a semi-automatic approach collecting face images from the Internet.
LFW~\cite{LFWTech} contains 13,233 faces of 5,749 identities. We follow the same protocol to preprocess them and 6,000 pairs (including 3,000 positive and 3,000 negative pairs) are selected to evaluate the face verification performance of the recognition models.
Age-DB~\cite{moschoglou2017agedb} contains 16,488 collected faces of 568 distinct subjects annotated with year, noise-free labels. Also, 6,000 pairs are selected.
UCCS ~\cite{uccs} contains 16,149 faces in 1,732 subjects, which is collected in real surveillance scenarios and is a benchmark for challenging face recognition in unconstrained scenario.
These faces are resized according to the student and teacher model input size accordingly.

Face verification task confirms whether two faces belong to  the same person using a 1:1 comparison. We extract features of each input pair and the accuracy is determined using constructed probe and gallery set pairs following the LFW protocol by calculating the cosine similarity with a pre-set threshold. 
Face identification attempts to identify the face by comparing it to gallery faces through a 1:N comparison.
In the experiment, the backbone network are fixed and the final layer are fine-tuned accordingly.
We compare the proposed approach following the standard top-K error metric.

In the experiments, we take the pretrained ArcFace~\cite{deng2019arcface} as the teacher which uses ResNet50 with a $112 \times 112$ input resolution. And ResNet18, a smaller model, as the student. Following previous work, we apply bicubic interpolation to obtain low-resolution faces with three resolutions of 16$\times$16, 32$\times$32 and 64$\times$64. The batch size is set as 96, and the learning rate is initialized as 0.05 and multiplied by 0.1 when the epoch is equal to 21, 28 and 32. All the experiments are implemented by PyTorch on a TITAN XP GPU.

\begin{figure*}[]
	\centering
	\vspace{-0.15in}
	\begin{minipage}{1\linewidth}	
        \subfigure[ArcFace\_LR-LR]{
			\label{fig:1}
			\includegraphics[width=0.32\linewidth,height=1in]{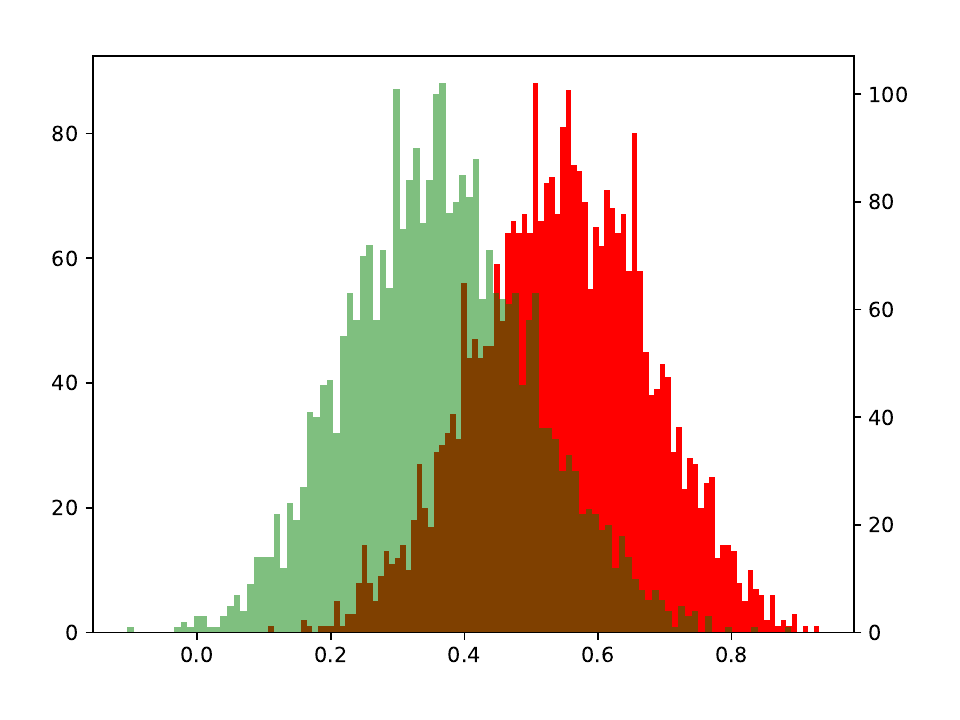}	
		}\noindent
		\subfigure[EKD\_LR-LR]{
			\label{fig:2}
			\includegraphics[width=0.32\linewidth,height=1in]{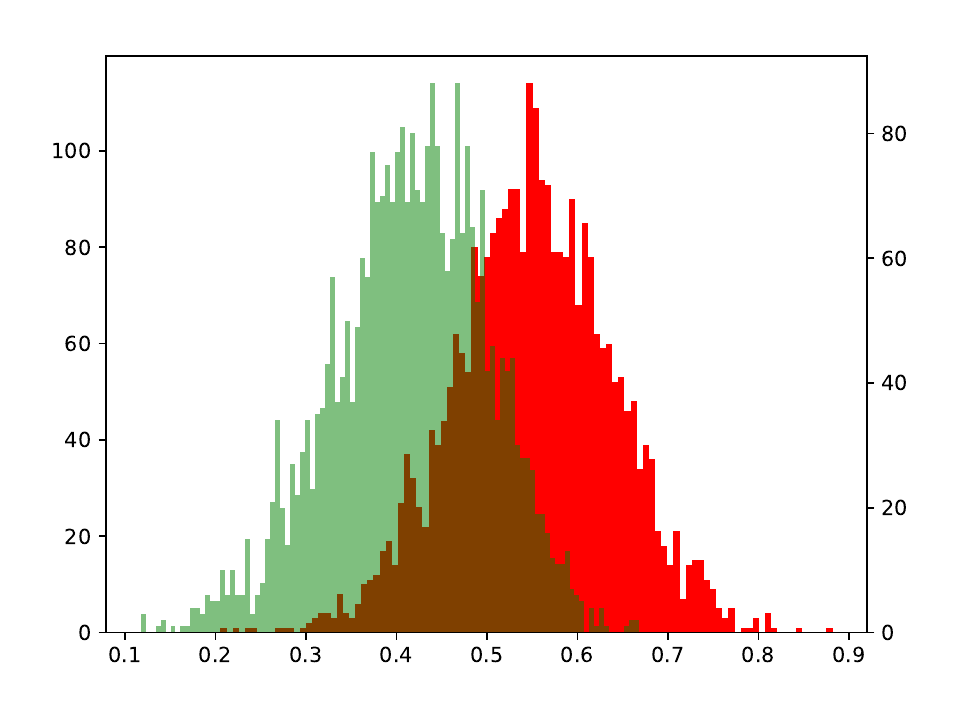}
		}\noindent
		\subfigure[AIRD\_LR-LR]{
			\label{fig:2-2}
			\includegraphics[width=0.32\linewidth,height=1in]{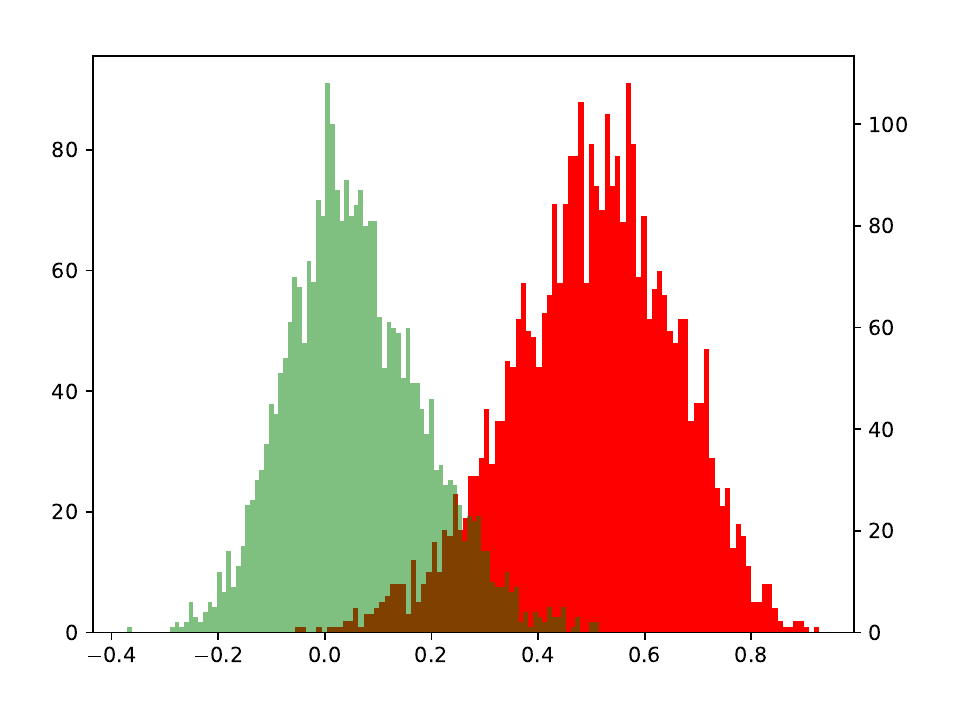}
		}
	\end{minipage}
	\vskip -0.3cm 
	\begin{minipage}{1\linewidth }
       \subfigure[ArcFace\_LR-HR]{
			\label{fig:3-2}
			\includegraphics[width=0.32\linewidth,height=1in]{{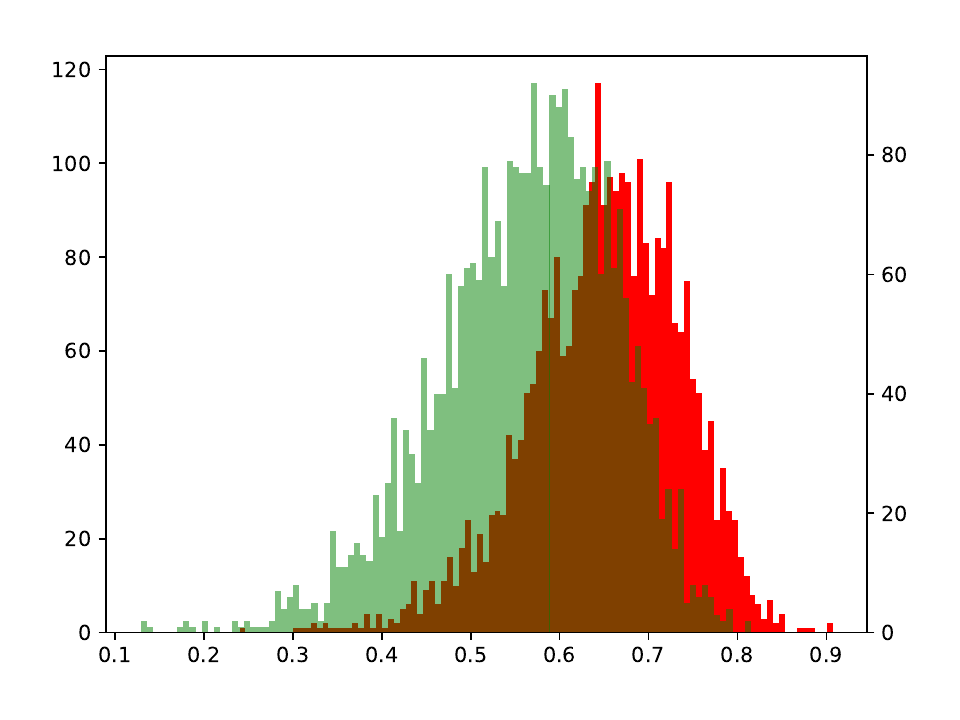}
			}
		}\noindent
		\subfigure[EKD\_LR-HR]{
			\label{fig:4-2}
			\includegraphics[width=0.32\linewidth,height=1in]{{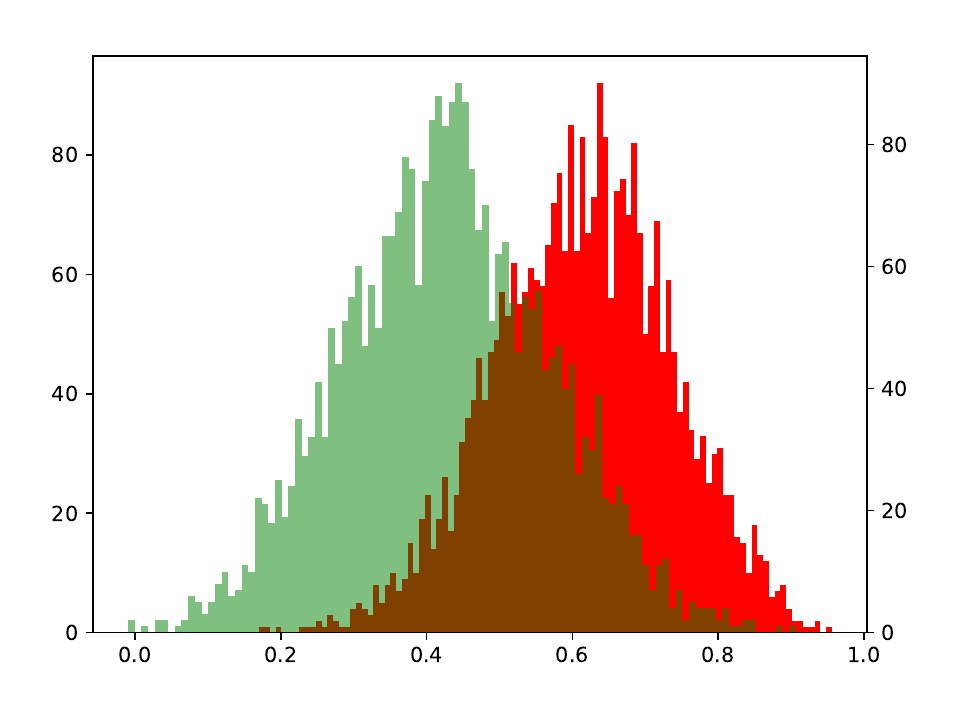}
			}
        }\noindent
		\subfigure[AIRD\_LR-HR]{
			\label{fig:4-3}
			\includegraphics[width=0.32\linewidth,height=1in]{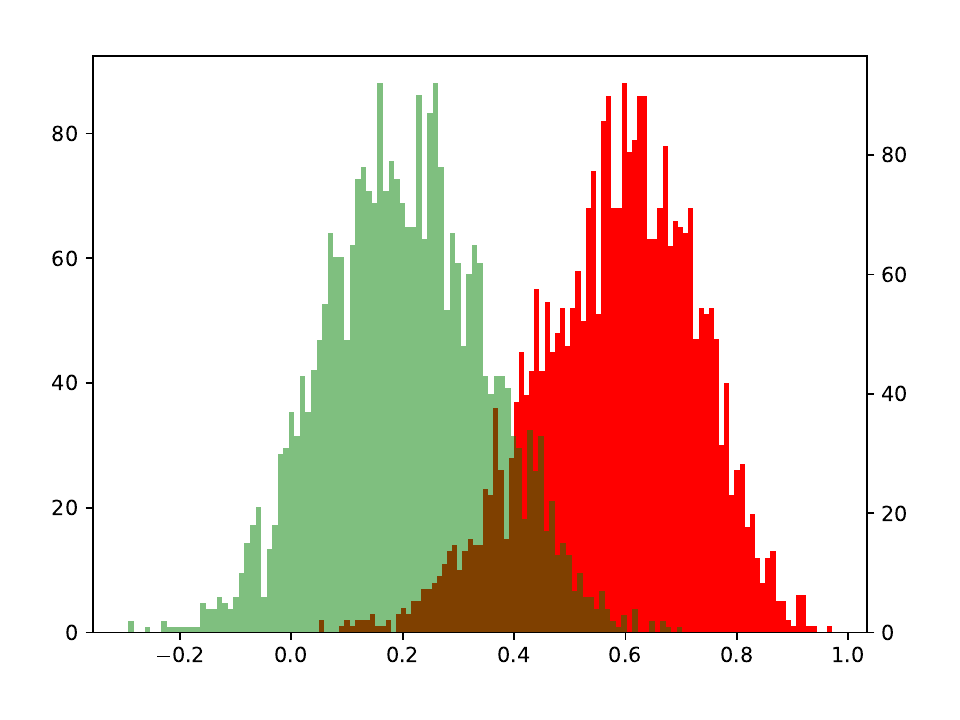}
		}
	\end{minipage}
	\vspace{-0.18in}	
	\caption{The score distributions achieved by different models on LFW.}
	\vspace{-0.2in}		
	\label{lfw-score}
\end{figure*}

\begin{table}[]
\caption{Comparison to baselines of face verification accuracy (\%) on LFW under LR-LR and LR-HR settings.}
\centering
\begin{tabular}{ccccc}
\hline
\multicolumn{1}{c|}{Scenarios}                       & \multicolumn{1}{c|}{Models}      & \multicolumn{1}{c}{LR=16 $\times$ 16} & \multicolumn{1}{c}{LR=32 $\times$ 32} & \multicolumn{1}{c}{LR=64 $\times$ 64} \\ \hline
\multicolumn{1}{c|}{\multirow{7}{*}{LR-LR}} & \multicolumn{1}{c|}{ArcFace\_I}  & 84.10                      & 95.62                     & 98.33                     \\
\multicolumn{1}{c|}{}                       & \multicolumn{1}{c|}{ArcFace\_II} & 92.30                      & 97.45                     & 98.95                     \\
\multicolumn{1}{c|}{}                       & \multicolumn{1}{c|}{CosFace\_I}  & 85.85                     & 94.33                     & 98.50                      \\
\multicolumn{1}{c|}{}                       & \multicolumn{1}{c|}{CosFace\_II} & 93.80                      & 97.13                     & 98.75                     \\
\multicolumn{1}{c|}{}                       & \multicolumn{1}{c|}{MegFace\_I}  & 88.65                     & 96.95                     & 98.62                     \\
\multicolumn{1}{c|}{}                       & \multicolumn{1}{c|}{MegFace\_II} & 94.97                     & 98.07                     & 99.15                     \\
\multicolumn{1}{c|}{}                       & \multicolumn{1}{c|}{\textbf{Our AIRD}}        & 96.02                     & 98.63                     & 99.20                      \\ \hline
\multicolumn{1}{c|}{\multirow{7}{*}{LR-HR}} & \multicolumn{1}{c|}{ArcFace\_I}  & 80.70                     & 88.26                     & 89.38                     \\
\multicolumn{1}{c|}{}                       & \multicolumn{1}{c|}{ArcFace\_II} & 67.17                     & 85.80                      & 94.62                     \\
\multicolumn{1}{c|}{}                       & \multicolumn{1}{c|}{CosFace\_I}  & 79.35                     & 87.45                     & 87.26                      \\
\multicolumn{1}{c|}{}                       & \multicolumn{1}{c|}{CosFace\_II} & 65.65                     & 84.38                     & 92.42                     \\
\multicolumn{1}{c|}{}                       & \multicolumn{1}{c|}{MegFace\_I}  & 82.71                     & 89.35                     &  88.95                   \\
\multicolumn{1}{c|}{}                       & \multicolumn{1}{c|}{MegFace\_II} & 72.75                     & 87.88                     & 93.85                     \\
\multicolumn{1}{c|}{}                       & \multicolumn{1}{c|}{\textbf{Our AIRD}}        & 91.52                     & 94.50                      & 99.45                     \\ \hline
\end{tabular}
\label{lfw-baseline}
\end{table}

\subsection{Results on Face Verification}
We consider two practical low-resolution face recognition scenarios where the probe face is low-resolution and the gallery face is low or high-resolution, denoting as LR-LR and LR-HR, respectively. We conduct the comparisons with state-of-the-art low-resolution models, including 4 normal-resolution ones(FaceNet~\cite{schroff2015facenet}, CosFace~\cite{wang2018cosface}, ArcFace~\cite{deng2019arcface} and MagFace~\cite{meng2021magface}) and 7 low-resolution models (SKD~\cite{ge2018low}, BridgeKD~\cite{ge2020efficient}, HORKD~\cite{GeHORKD20}, EKD~\cite{zhang2021student}, RPCL~\cite{li2022deep}).
To ensure fair comparisons, we follow the protocols as in previous studies to prepare the data for evaluation.
The accuracy is the percentage of correct predictions, where the threshold is decided as the one with the highest accuracy.

First, we consider two kinds of baselines. The first kind is directly using pre-trained teacher with high-resolution faces, which is denoted as model\_I. The second kind is the student trained with the low-resolution faces from scratch, denoted as model\_II.
These experiments are designed to demonstrate the validity of our distillation framework, as well as the effectiveness of knowledge transfer.
As in Tab~\ref{lfw-baseline}, three main conclusions can be drawn.
First, lower-resolution result in evident accuracy drop, which is in accord with discoveries of previous researches.
Second, LR-HR is more difficult than LR-LR scenario because of the resolution of the input differences significantly affects the accuracy. 
Finally, our AIRD outperforms two baselines in both scenarios, especially when probe resolution is lower, which is more challenging. 

\begin{table}[]
\centering
\caption{Face verification accuracy (\%) on LFW under two practical scenarios. Here, LR is 16$\times$16 and HR is normal resolution.}
\begin{tabular}{c|c|c}
\hline
 Models & LR-LR & LR-HR  \\ \hline
FaceNet~\cite{schroff2015facenet}~(CVPR 2015)                 & 90.25            & 65.33     \\ 
CosFace~\cite{wang2018cosface}~(CVPR 2018) & 93.80  & 67.17       \\ 
ArcFace~\cite{deng2019arcface}~(CVPR 2019) & 92.30   & 65.65   \\ \
MagFace~\cite{meng2021magface}~(CVPR 2021)   & 94.97      & 72.75    \\ \hline
SKD~\cite{ge2018low}~(TIP 2019)  & 85.87     &    --   \\
BridgeKD~\cite{ge2020efficient}~(TIP 2021)   & 85.88      & 71.30       \\ 
HORKD~\cite{GeHORKD20}~(AAAI 2020)      & 90.03     & 80.87   \\
EKD~\cite{zhang2021student}~(TCSVT 2022)  & 91.71    & 81.15       \\ 
RPCL-CosFace~\cite{li2022deep}~(NN 2022)     & 95.13        & 88.26     \\ 
RPCL-ArcFace~\cite{li2022deep}~(NN 2022)  & 94.70       & 89.33       \\ 
RPCL-MagFace~\cite{li2022deep}~(NN 2022)             & 95.12   & 90.62      \\ 
CRRCD~\cite{zhang2023low}~(TCSVT 2023)            & 95.25   & 91.33      \\ 
\textbf{Our AIRD}                  & \textbf{96.02}        & \textbf{91.52}     \\ \hline
\end{tabular}
\label{lfw}
\end{table}

\begin{table}[b]
\centering
\caption{Face verification accuracy (\%) on Age-DB under two practical scenarios. Here, LR is 16$\times$16 and HR is normal resolution.}
\begin{tabular}{c|c|c}
\hline
 Models & LR-LR & LR-HR  \\ \hline
ArcFace~\cite{deng2019arcface}~(CVPR 2019) & 77.75   & 52.83   \\ \
F-KD~\cite{massoli2020cross}~(IVC 2020)  & 77.85     &    53.05   \\
AT~\cite{komodakis2017paying}~(ICLR 2017)   & 77.40      & 60.72       \\ 
HORKD~\cite{GeHORKD20}~(AAAI 2020)      & 78.27     & 63.93   \\
EKD~\cite{zhang2021student}~(TCSVT 2022)  & 77.35    & 63.58       \\ 
A-SKD~\cite{shin2022teaching}~(ECCV 2022)            & 79.00   & 65.08      \\ 
\textbf{Our AIRD}                  & \textbf{80.45}        & \textbf{66.25}     \\ \hline
\end{tabular}
\label{Age-DB}
\end{table}
In order to verify the robustness of our low-resolution student models, we then emphatically check the accuracy when the input resolution is $16 \times 16$. We conduct comparisons with the state-of-the-art low-resolution face recognition methods at a much lower-resolution on both LFW~\cite{LFWTech} and Age-DB~\cite{moschoglou2017agedb}, as in Tab~\ref{lfw} and Tab~\ref{Age-DB}.
Compared to distillation-based ones, AIRD achieves better accuracy since it extracts instance-relation level knowledge which is more effective than lower-order transferring.
Also, our AIRD performs well in both LR-LR and LR-HR scenarios, demonstrating the robustness of our method.
RPCL~\cite{li2022deep} models learn margin-based discriminative features, our method still outperforms it in both recognition scenarios since it implicitly encodes
the knowledge by using anchor-based high-order relation preserving distillation.

To provide a more intuitional demonstration, Fig.~\ref{lfw-score} exhibits the distribution of similarity scores achieved by different models on LFW under LR-LR and LR-HR settings. 
In each subfigure, the scores of the positive and negative pairs are presented in green and red color, separately. Smaller overlapping suggests a more distinct separation between pairs. It is clear that the proposed AIRD presents a smaller overlapping.
Compared with the others, our method shows better performance.

\subsection{Results on Face Identification}
UCCS is collected in real surveillance scenarios, including blurred images and bad illumination, which is a challenging benchmark for face recognition in unconstrained scenarios. We follow the training and testing settings in~\cite{ge2020efficient, GeHORKD20}, randomly select a 180-subject subset, separate the faces into 3918 training and 907 testing faces, and finetune the layer parameters on training set.
We show the accuracy at a resolution of $16 \times 16$.

As shown in Tab.~\ref{uccs}, AIRD achieves the best identification accuracy compared with others, at least an improvement of 1.56\%.
This suggests that although low-resolution faces lack some important information needed for recognition, our ARID can learn discriminative representations that are beneficial for improving low-resolution face recognition performance.
Classical methods, such as ArcFace, do not perform well, with their highest recognition accuracy only reaching 88.73\%.
Moreover, our method still obtains higher accuracy compared with distillation-based methods, which means that our method can help the student model learn richer feature representations.

\begin{figure}[t]
	\centering
	\vspace{-0.15in}
	\begin{minipage}{1\linewidth}	
		\subfigure[VLRR]{
			\label{fig:1-1}
			\includegraphics[width=0.46\linewidth,height=1.4in]{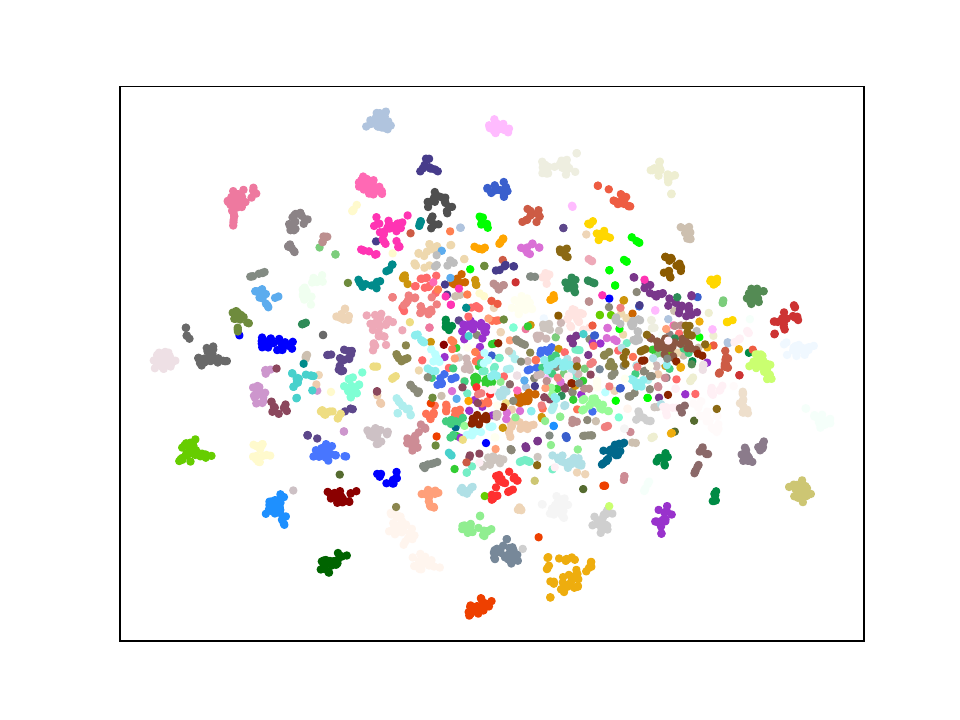}	
		}\noindent
		\subfigure[Arcface]{
			\label{fig:2-1}
			\includegraphics[width=0.46\linewidth,height=1.4in]{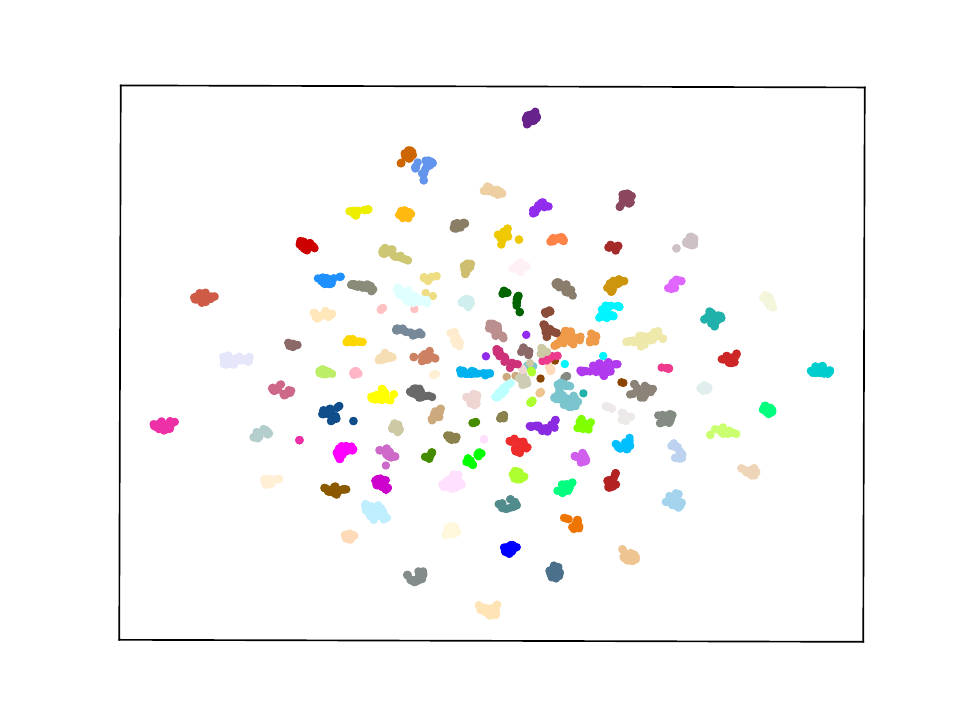}
		}
	\end{minipage}
	\vskip -0.3cm 
	\begin{minipage}{1\linewidth }
		\subfigure[EKD]{
			\label{fig:3}
			\includegraphics[width=0.46\linewidth,height=1.4in]{{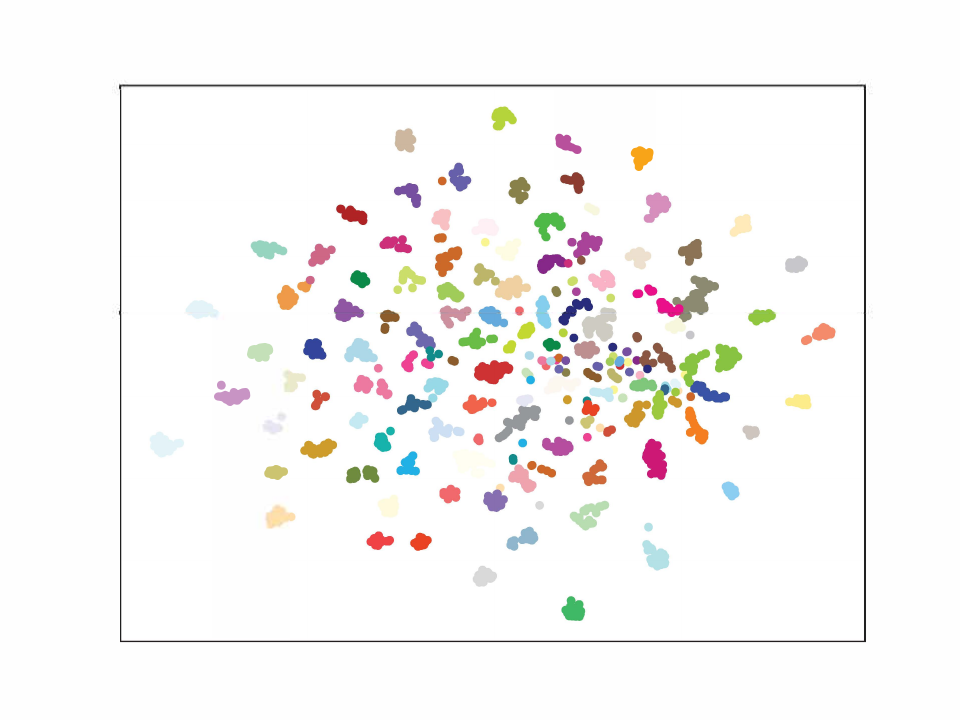}
			}
		}\noindent
		\subfigure[AIRD]{
			\label{fig:4}
			\includegraphics[width=0.46\linewidth,height=1.4in]{{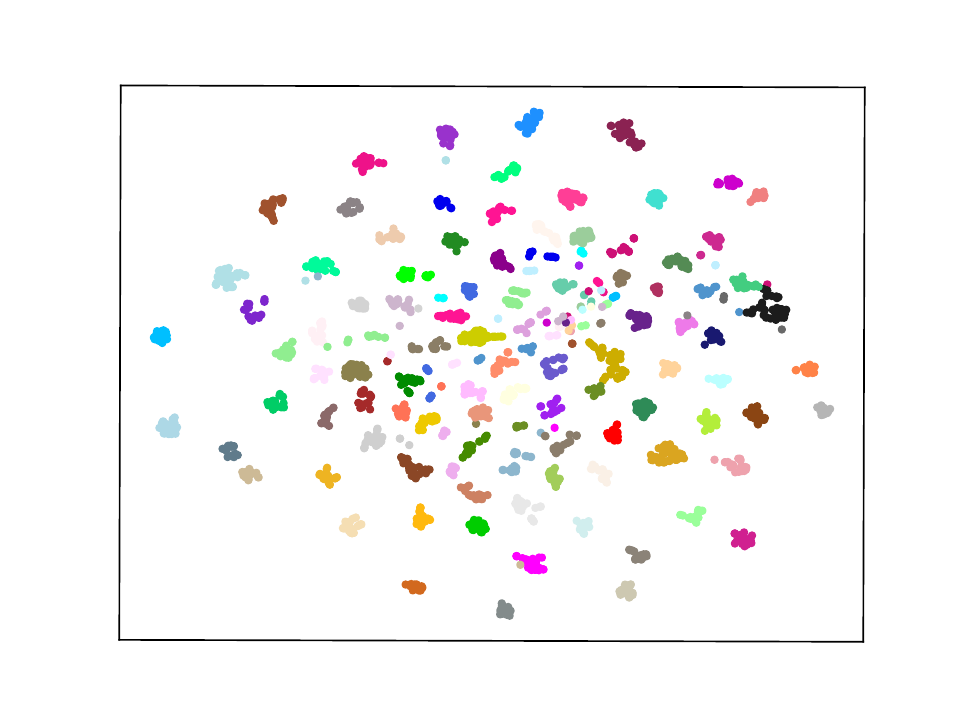}
			}
		}
	\end{minipage}
	\vspace{-0.18in}	
	\caption{t-SNE feature plots on UCCS with four models.}
	\vspace{-0.2in}		
	\label{tsne-uccs}
\end{figure}

To further demonstrate the advantages of AIRD, we use t-SNE for visualization. 
It converts similarities between data points to joint probabilities and tries to minimize the KL divergence between the joint probabilities of the low-dimensional embedding and the high-dimensional data. 
We randomly select 100 identities from UCCS, different color indicate different identities in Fig.~\ref{tsne-uccs}. It is obvious that our approach achieves more concentrated clusters than baseline.

\begin{table}[t]
	\centering
	\caption{Face identification performance on UCCS benchmark. All approaches work at a low resolution of $ 16\times16 $, and our student outpeforms 12 SOTAs by at least an accurracy improvement of 1.46\%.}
	\renewcommand{\arraystretch}{1.1}
	\begin{tabular}{c|c}
		\hline
		Models & Accuracy(\%) \\
		\hline
		VLRR~\cite{wang2016studying}~(CVPR 2016)    & 59.03   \\
		SphereFace~\cite{liu2017sphereface}~(CVPR 2017) & 78.73  \\
		CosFace~\cite{wang2018cosface}~(CVPR 2018) & 91.83  \\
		SKD~\cite{ge2018low}~(TIP 2019)   & 67.25  \\
		AGC-GAN~\cite{talreja2019attribute}~(BTAS 2019)   & 70.68  \\
		VGGFace2~\cite{Cao2018VGGFace2}~(FG 2018)   & 84.56  \\
		ArcFace~\cite{deng2019arcface}~(CVPR 2019)  & 88.73  \\
		HORKD~\cite{GeHORKD20}~(AAAI 2020)   & 92.11  \\
		LRFRW~\cite{li2019low}~(CTIFS 2019)   & 93.40  \\
		CSRIP~\cite{grm2019face}~(TIP 2019) & 93.49 \\
		EKD~\cite{zhang2021student}~(TCSVT 2022) & 93.85  \\
		DirectCapsNet~\cite{singh2019dual}~(ICCV 2019) & 95.81 \\
        CRRCD~\cite{zhang2023low}~(TCSVT 2023)            & 97.27   \\ 
		\textbf{Our ARID} & \textbf{97.58}  \\
		\hline
	\end{tabular}
	\label{uccs}
\end{table}

\subsection{Ablation Studies}
Ablation studies are conducted to validate the effectiveness of each
components on LFW and Age-DB under LR-LR settings with the resolution of 16 $\times$ 16 to evaluate the efficiency of the detailed design of our method.
We conduct experiments to show the contribution of each loss component. 
Our ARID consists of instance-relation distillation and inference adaptation.

First, we show the effect of instance-relation distillation without FaceBN.
As in Tab.~\ref{tab}, both IlD and RlD improve the results compared to baseline which only use a general classification loss for face recognition. Besides, RlD increases more, which means that transferring high-order relational contrastive knowledge is helpful for student to learn discriminative representations.
When combining them, AIRD achieves a higher accuracy, which means that only instance or relation level knowledge transfer may not enough and these two can complement each other.
This illustrates the effectiveness of our instance-relation distillation.

\begin{figure}[htbp]
    \centering 
    \includegraphics[width=0.49\linewidth,height=1.4in]{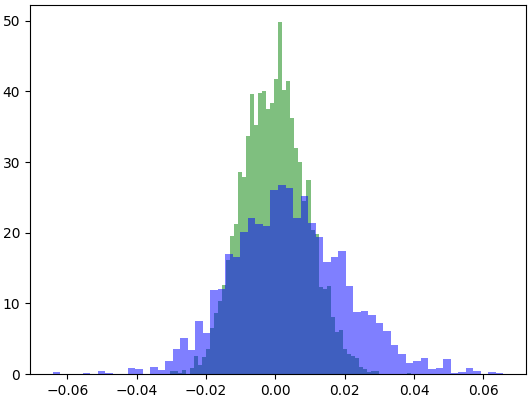}
    \includegraphics[width=0.49\linewidth,height=1.4in]{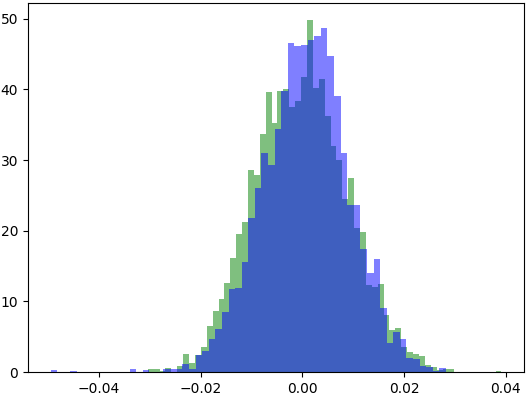}
    \caption{Data distribution statistics before (left) and after (right) FaceBN of LFW and CASIA-WebFace. }
    \label{distribution}
\end{figure}

Second, as in Tab.~\ref{tab}, applying FaceBN further improves the accuracy, demonstrating the adaptability of FaceBN. 
Moreover, we show the data mean distribution of LFW (in blue) and CASIA-WebFace (in green) in Fig.~\ref{distribution} visually. 
We train the student model with the CASIA-WebFace dataset, and observe the distribution statistics in the feature map during inference using LFW.
Fig.~\ref{distribution} shows the close-ups. The feature distribution of LFW with the traditional batch normalization has an obvious shift compared with that of CASIA-WebFace, which is the training dataset. Fortunately, equipped with FaceBN, the distribution statistics of LFW seem more close to the training dataset. 
This indicates that FaceBN can mitigate data gap, delivering a reasonable distribution
to the following layers. In this way, it can improve the generalization ability of models and facilitate recognition accuracy.

\begin{table}[htbp]
\centering
\caption{The component effect on verification accuracy (\%).} 
\begin{tabular}{ccccccc}
\cline{1-6}
$\mathcal{L}_{Cls}$    & $\mathcal{L}_{IlD}$    & \multicolumn{1}{c|}{$\mathcal{L}_{RlD}$} & \multicolumn{1}{c|}{FaceBN} & \multicolumn{1}{c}{LFW} & Age-DB \\ \cline{1-6}
$\checkmark$  &  - & \multicolumn{1}{c|}{-} & \multicolumn{1}{c|}{-}  & \multicolumn{1}{c}{92.30}  & 77.75   \\
$\checkmark$  & $\checkmark$   & \multicolumn{1}{c|}{-}  & \multicolumn{1}{c|}{-}  & \multicolumn{1}{c}{93.83}  & 78.05  \\
$\checkmark$  & -  & \multicolumn{1}{c|}{$\checkmark$ }     & \multicolumn{1}{c|}{-}           & \multicolumn{1}{c}{95.25}  & 79.72    \\
$\checkmark$                     & $\checkmark$                     & \multicolumn{1}{c|}{$\checkmark$}     & \multicolumn{1}{c|}{-}   & \multicolumn{1}{c}{95.76}  & 79.85                  \\
$\checkmark$                     & $\checkmark$                     & \multicolumn{1}{c|}{$\checkmark$}     & \multicolumn{1}{c|}{$\checkmark$}          & \multicolumn{1}{c}{96.02}  & 80.45                \\ \cline{1-6}
\label{tab}
\end{tabular}
\end{table}

\begin{figure}[htbp]
    \centering 
    \centering
\centerline{\includegraphics[width=7cm,height=5cm]{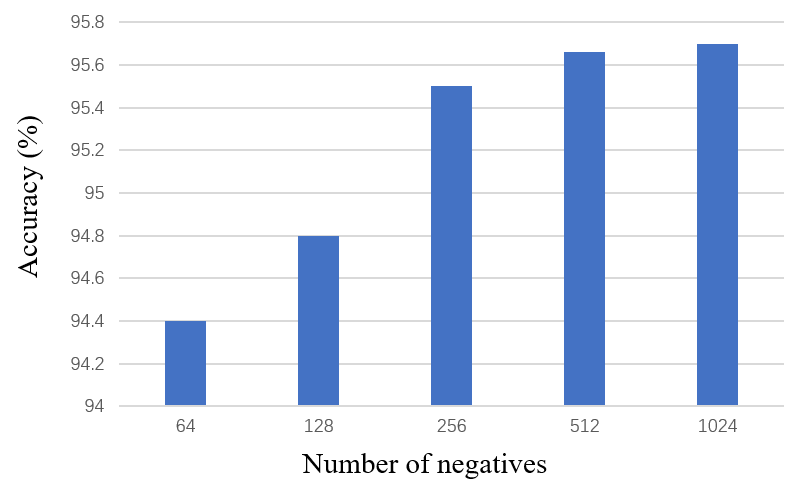}}
    \caption{Face verification accuracy on LFW under different negative number. }
    \label{sample}
\end{figure}

Third, we verify the effect of pre-selecting positive and negative pairs' sampling policy, which is one of the main differences of the previous, like CRCD~\cite{zhu2021complementary} and DKD~\cite{zhao2022dkd}.
We consider our sampling method as well as random sampling. As in Fig.~\ref{sample}, pre-selecting positive and negative pairs can bring at least 0.25\% improvement at accuracy on LFW.
Then, we also validate five different negative pairs numbers (64, 128, 256, 512 and 1024), as the negative number has a crucial impact on the final performance in contrastive learning. The results is in the right of Fig.~\ref{sample}. Here, increasing negative number will lead to performance improvement, which means
higher-order relation knowledge is built. However, larger negative number requires more computations, which means the negative number should be carefully selected to balance the accuracy and computation cost.

\section{CONCLUSION}
In this paper, we propose an adaptable instance-relation distillation approach to facilitate low-resolution face recognition. The approach distills the teacher knowledge from instance and relation levels, and then the learned student model can be used to effectively recognize testing low-resolution faces via inference adaptation. 
In this way, our method improves the transfer ability of the learned model in recognizing real-world low-resolution faces.
Extensive experiments on low-resolution face recognition tasks show the effectiveness and adaptability of our proposed approach. 
In the future, we will explore its potential on more low-resolution object recognition tasks.

\bibliographystyle{IEEEtran}
\bibliography{IEEEexample}

\end{document}